\documentclass[10pt,twocolumn,letterpaper]{article}

\usepackage{abc}
\usepackage{times}
\usepackage{epsfig}
\usepackage{graphicx}
\usepackage{amsmath}
\usepackage{amssymb}
\usepackage{url}

\abcfinalcopy

\ifabcfinal\fi
\begin{document}

\title{MobiFace: A Lightweight Deep Learning Face Recognition on Mobile Devices}

\author{Chi Nhan Duong $^{1}$, Kha Gia Quach $^{1}$, Ibsa Jalata $^{3}$, Ngan Le $^{2}$, Khoa Luu $^{3}$ \\
	$^{1}$ Computer Science and Software Engineering, Concordia University, Canada\\
	$^{2}$ Electrical and Computer Engineering, Carnegie Mellon University, USA\\
	$^{3}$ Computer Science and Computer Engineering, University of Arkansas, USA\\
	\tt\small $^{1}$\{dcnhan, kquach\}@ieee.org,  $^{2}$thihoanl@andrew.cmu.edu,\\ \tt\small $^{3}$\{ikjalata, khoaluu\}@uark.edu
}

\maketitle
\thispagestyle{empty}

\begin{abstract}
Deep neural networks have been widely used in numerous computer vision applications, particularly in face recognition. However, deploying deep neural network face recognition on mobile devices has recently become a trend but still limited since most high-accuracy deep models are both time and GPU consumption in the inference stage.
Therefore, developing a lightweight deep neural network is one of the most practical solutions to deploy face recognition on mobile devices. Such the lightweight deep neural network requires efficient memory with small number of weights representation and low cost operators.
In this paper, a novel deep neural network named MobiFace, a simple but effective approach, is proposed for productively deploying face recognition on mobile devices.
The experimental results have shown that our lightweight MobiFace is able to achieve  high performance with 99.73\% on LFW database and 91.3\% on large-scale challenging Megaface database. It is also eventually competitive against large-scale deep-networks face recognition while significant reducing computational time and memory consumption.
\end{abstract}

\section{Introduction}

Deep Convolutional Neural Network (CNNs) have revolutionized numerous machine learning applications with high accuracy, even in some applications, the performance exceeds human level of accuracy. Most of the computer vision applications, e.g. object detection \cite{girshick2014rcnn,Luu_FD_CVPR2018,Luu_FD_FG2018,Luu_FR_ICCV2017, Zheng_FD_BTAS2016, Zhu_FD_CVPRW2016, Le_FD_ICPR2016}, object classification \cite{Kaiming-Resnet, Luu_FR_CVPR2016,Luu_FR_CVPR2015},
object segmentation \cite{he2017maskrcnn,Long-FCN,Luu_Seg_PR2018,Luu_Seg_TIP2018, le_miccai_2018}, and modeling \cite{Duong_IJCV2018} etc., deploy CNNs with other Deep Neural Network framework to achieve the highest accurate performance. However, using deeper neural network with hundreds of layer and millions of parameters to achieve higher accuracy comes at cost. The networks require high computational resources beyond the capabilities of many mobile and embedded applications. 
To deploy such networks and to get higher performance, powerful GPUs with larger memory size are needed. As a result of this limitation, recently, there is an advancement in compressing deep neural networks known as compressed networks. Some familiar compressed networks are Mimic Networks \cite{Li2017MimickingVE,Wei2018QuantizationMT}, Pruning \cite{Han-LBWs,Han-Deep,Liu2017LearningEC}, Depth-wise Convolution \cite{Howard-MobileNet,Sandler-MNetv2}, BinaryNets \cite{Itay-BNN,Matt-BinaryNet,Rastegari-xnor,CourbariauxBD15}. These networks improve the speedup of the inference stage without significant lose of accuracy. Meanwhile, the improvements have not benchmarked in face recognition like in image classification and/or object detection. Moreover, face recognition problems are expected to be robust such as the very deep features of millions of facial subjects discriminated. In this problem, a noticeable number of neural layers should be included in the model unlike detection or classification.

In this work, our main contribution is a novel lightweight but high performance deep neural network for face recognition on mobile devices. Our network highly minimizes the number of operations and memory required while retaining the same accuracy mobile tailored computer vision models. In contrast to previous work, we present our contributions in MobiFace approach as follows: (1) we improve the well-known framework MobileFaceNet \cite{chen2018MobileFaceNets} by changing to further lighter-weight with higher accuracy performance. (2) The proposed method MobiNet is then applied in face recognition and optimized within an end-to-end deep learning framework. In addition, we extensively experiment our proposed MobiNet on both mobile-based network and large-scale deep-network on face recognition tasks with two state-of-the-art face recognition databases,i.e. Labeled Faces in the Wild (LFW) and large-scale challenging Megaface databases.

\section{Prior Work}

There has been a rising interest in lightweight deep network design mainly in tuning deep neural architectures to strike an optimal balance between accuracy and performance for the past decade. There are many lightweight deep networks that can be categorized into designed compact modules and pruned networks, binarized networks, quantized networks and mimicked networks. Our mainly focus in this section will be the first two categories where our proposed framework mainly related with.

In MobileNet \cite{Howard-MobileNet}, Andrew et al. proposed depthwise separable convolutions to build light weight deep neural networks. In this work, the standard depthwise convolution is factorized into a depthwise convolution where a convolution applies to a single filter for each channel and a pointwise convolution where $1\times1$ convolution applied to combine the output from the depthwise convolution. VGG-16 with 138 million parameters and MobileNet with 4.2 million parameters achieve about the same accuracy on ImageNet \cite{imagenet_cvpr09}. Sandler et al. in \cite{Sandler-MNetv2} known as MobileNet-V2 further improves MobileNet on different benchmarks and multiple tasks. This work presents inverted residuals and linear bottlenecks \cite{Kaiming-Resnet} based on an inverted residual structure where the shortcut connections are between the thin bottleneck layers. In addition, the intermediate expansion layer uses lightweight depthwise convolutions to filter features as a source of non-linearity. It slightly improves the performance of MobileNet \cite{Howard-MobileNet} where the accuracy is 70.6\% to 72\% on ImageNet \cite{imagenet_cvpr09} with only 3.4 million parameters. To reduce the memory and computational cost, FD-Mobilenet \cite{qin2018fd} and MobileFaceNets  \cite{chen2018MobileFaceNets} were presented based on MobileNet-V2 framework but still there are so much work to be done to effectively run deep learning framework on CPU.

Along side of compression methods, Han et al. \cite{Han-LBWs}  proposed to prune trivial connections by an absolute value without losing accuracy performance. Liu et al. \cite{Liu2017LearningEC} proposed a novel learning scheme to slim a network. In their work, instead of using absolute values of weighs, they impose sparsity induced regularization on the scaling factor in Batch Normalization to slim the network and thus insignificant channels can be spotted. Overall, slimming a network proved to be successful in producing better results in ResNet \cite{Kaiming-Resnet}, DenseNet \cite{Gao-Dense} and VGG-16 \cite{Simonyan-VGGNet}. However, for each pruned connection, a list of indices needs to be stored to memory, leading to a very low progress for both training and testing.

\section{Our Proposed MobiNet}

This section starts with introducing network design strategies to construct a lightweight deep network. Then, by adopting these strategies, the architecture of MobiFace for face recognition on mobile devices is introduced. Thanks to the concise and precise deep network architecture, the proposed framework is efficient in terms of small computational cost and high accuracy in comparison against other deep networks on several large-scale face recognition databases.

\subsection{Network Design Strategy}
\textbf{Bottleneck Residual block with the expansion layers}. The use of  Bottleneck Residual block is introduced in \cite{sandler2018mobilenetv2} where a block consists of three main transformation operators, i.e. two linear transformations and one non-linear per-channel transformation.
There are three key factors of this type of block: (1) the non-linear transformation to learn complex mapping functions; (2) the layer expansion with increasing number of feature maps in the inner layers; and (3) shortcut connections to learn the residual. 
Formally, given an input $\mathbf{x}$ with the size of $h \times w \times k$, a bottleneck residual block can be represented as follows,
\begin{equation}
 \mathcal{B}(\mathbf{x}) = [\mathcal{F}_1 \circ \mathcal{F}_2 \circ \mathcal{F}_3] (\mathbf{x})   
\end{equation}
where $\mathcal{F}_1:\mathbb{R}^{w \times h \times k} \mapsto \mathbb{R}^{w \times h \times tk}$ and $\mathcal{F}_3:\mathbb{R}^{w \times h \times k} \mapsto \mathbb{R}^{\frac{w}{s} \times \frac{h}{s} \times k_1}$ are the linear function represented by $1 \times 1$ convolution operator, and $t$ denotes the expansion factor. $\mathcal{F}_2:\mathbb{R}^{w \times h \times tk} \mapsto \mathbb{R}^{\frac{w}{s} \times \frac{h}{s} \times tk}$ is the non-linear mapping function which is a composition of three operators, i.e. ReLU, $3 \times 3$ depthwise convolution with stride $s$, and ReLU.

The residual learning connection is employed in a bottleneck block.
This type of blocks is shown to have the capabilities of preventing manifold collapse during transformation and also increasing the expressiveness of the feature embedding \cite{sandler2018mobilenetv2}.

\textbf{Fast Downsampling}. Under the limited computational resource of mobile devices, a compact network should \textit{maximize the information transferred from the input image to output features} while \textit{avoiding the high computational cost due to the large spatial dimensions} of feature maps. 
In the large-scale deep networks, the detail information flow is usually ensured by the \textit{slow downsampling strategy}, i.e. the spatial dimensions are slowly reduced between blocks by downsampling operator. Consequently, the these networks maintain so many feature maps with large spatial size and result in a heavy-size network.
On the other hand, under the limited computational budgets, a light-weight network adopting that slow downsampling may suffer both issues of weak feature embedding and high processing time. 

In these cases, \textit{fast downsampling strategy} can be considered as an efficient replacement of the slow downsampling technique. In particular, in fast downsampling, the downsampling steps are consecutively applied in the very beginning stage of the feature embedding process to avoid large spatial dimension of the feature maps. Then in the later stage, more feature maps are added to support the information flow of the whole network. By this way, more complex mapping functions are learned to generate more details feature. Notice that, in this strategy, even more feature maps were added to the later feature, i.e. increase the number of channels, the computational cost is maintained to be low since the spatial dimensions of these feature maps are small.

\begin{table}[!t]
	\small
	\centering
	\caption{Model Architecture for facial feature embedding. $/2$ means the operator has stride of 2. ``Block'' and ``RBlock'' indicate the Bottleneck Block and Residual Bottleneck block, respectively.}
	\label{tb:network_structure} 
	\begin{tabular}{|c|l|}
		\hline
		\textbf{Input} & \textbf{Operator} \\  \hline \hline
		112 $\times$ 112 $\times$ 3 & 3 $\times$ 3 Conv, $/2$, 64\\
		\hline
		56 $\times$ 56 $\times$ 64 & 3 $\times$ 3 DWconv, 64\\
		\hline
		56 $\times$ 56 $\times$ 64 &  Block  1 $\times$
		                      \Bigg\{
                              \begin{tabular}{lll}
                              1 $\times$ 1 Conv, 128 \\
                              3 $\times$ 3 DWconv, $/2$, 128 \\
                              1 $\times$ 1 Conv, Linear, 64 
                              \end{tabular}
                             \\
		\hline
		28 $\times$ 28 $\times$ 64 & RBlock 2 $\times$
		                      \Bigg\{
                              \begin{tabular}{lll}
                              1 $\times$ 1 Conv, 128 \\
                              3 $\times$ 3 DWconv, 128 \\
                              1 $\times$ 1 Conv, Linear, 64 
                              \end{tabular}
                            \\
        \hline
		28 $\times$ 28 $\times$ 64 & Block  1 $\times$
		                      \Bigg\{
                              \begin{tabular}{lll}
                              1 $\times$ 1 Conv, 256 \\
                              3 $\times$ 3 DWconv, $/2$, 256 \\
                              1 $\times$ 1 Conv, Linear, 128 
                              \end{tabular}
                            \\
        \hline
		14 $\times$ 14 $\times$ 128 & RBlock 3 $\times$
		                      \Bigg\{
                              \begin{tabular}{lll}
                              1 $\times$ 1 Conv, 256 \\
                              3 $\times$ 3 DWconv, 256 \\
                              1 $\times$ 1 Conv, Linear, 128 
                              \end{tabular}
                            \\
        \hline
		14 $\times$ 14 $\times$ 128 & Block 1 $\times$
		                      \Bigg\{
                              \begin{tabular}{lll}
                              1 $\times$ 1 Conv, 512 \\
                              3 $\times$ 3 DWconv, $/2$, 512 \\
                              1 $\times$ 1 Conv, Linear, 256 
                              \end{tabular}
                            \\
        \hline
		7 $\times$ 7 $\times$ 256 & RBlock 6 $\times$
		                      \Bigg\{
                              \begin{tabular}{lll}
                              1 $\times$ 1 Conv, 512 \\
                              3 $\times$ 3 DWconv, 512 \\
                              1 $\times$ 1 Conv, Linear, 256 
                              \end{tabular}
                            \\
        \hline
        7 $\times$ 7 $\times$ 256 & 1 $\times$ 1 Conv, 512\\
        \hline
        7 $\times$ 7 $\times$ 512 & 512-d FC\\
        \hline
	\end{tabular}
\end{table}

\subsection{MobiFace}
In this section, we present a novel MobiFace approach, simple but efficient deep network for face recognition. 
Given an input facial image with the size of $112 \times 112 \times 3$, this light-weight network aims at maximizing the information embedded in final feature vector while maintaining the low computational cost.
Inspired by the strategies presented in the previous section, the Residual Bottleneck block with expansion layers is adopt as the building block of MobiFace. 
Table \ref{tb:network_structure} represents the main architecture of MobiFace that consists of one $3 \times 3$ convolutional layer,  one $3 \times 3$ depthwise separable convolutional layer, followed by a sequence of Bottleneck blocks and Residual Bottleneck blocks, one $1 \times 1$ convolutional layer, and a fully connected layer. 
The structures of Residual Bottleneck blocks and Bottleneck blocks are very similar except a shortcut is added to connect the input and the output of the $1 \times 1$ convolution layer. Moreover, the stride $s$ is set to $2$ in Bottleneck blocks while that parameter is set to $1$ in every layers of Residual Bottleneck blocks.

Moreover, we adopt the fast downsampling strategy in our network architecture by quickly reducing the spatial dimensions of layers/blocks with the input size larger than $14 \times 14$. As one can easily see that given an input image with the size of $112 \times 112 \times 3$, the spatial dimension is reduced by half within the first two layers and become $8\times$ smaller after the other $7$ bottleneck blocks. The expansion factor is kept to $2$ whereas the number of channels is double after each Bottleneck block in later feature embedding stage.

A batch normalization together with a non-linear activation are applied after each convolutional layer except the one marked as ``Linear''. In our implementation, PReLU is used for the non-linear activation function due to its accuracy improvement over ReLU function.
In the last layer of MobiFace, rather than employing the Global Average Pooling (GAP) layer as in previous approaches \cite{howard2017mobilenets,sandler2018mobilenetv2,chen2018MobileFaceNets}, we use the Fully Connected (FC) layer in the last stage of embedding process. Compared to GAP which treats very units in the last convolutional layer equally (\textit{which is not very efficient since the information in the center pixel should play more important role than the one in the corner of the input}), the FC layer can learn different weights to these units and gain the information embedded in the final feature vector.

\begin{figure}
	\centering
	\includegraphics[width=1\columnwidth]{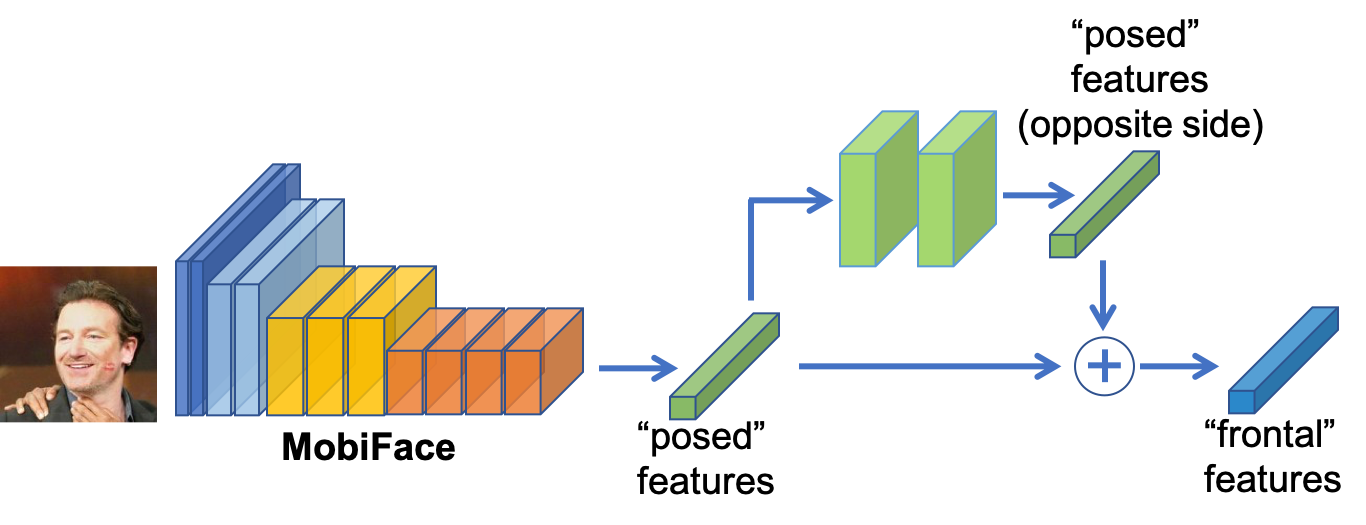}
	\caption{\textbf{The structure of the proposed Flipped-MobiFace}. The Flipped-MobiFace is an extended structure of MobiFace with two additional layers that learn to predict the deep features of the posed face in the opposite side.}
	\label{fig:FlippedMobiFace}
\end{figure}

\subsection{Flipped-MobiFace}

Now we describe a simple but computationally efficient approach to further improve the performance of our proposed MobiFace against pose variation. It is worth noting that during learning process of MobiFace, the deeper layers tend to extract deep features that favours the linear separability between classes to support the linearization process in latent domain. In other words, after embedding the faces of different subjects from images to feature domain, they become linear separable. This linearity property of deep feature domain inspires us to a hypothesis that the averaging of the deep features of left- and right-posed faces within the same angle can approximate the deep features of the frontal face. Thus, given a face at a particular pose, by simply flipping the input image (i.e. approximate the face of the opposite side) and computing their average deep features, the effects of pose variation can be miminized in this new features.
However, the extra computation will occur when extract features of the mirror image in addition to the original image. Therefore, we proposed the flipped version of MobiFace, namely Flipped-MobiFace, that requires to run feature extraction network once. By this way, it can be efficiently deployed on mobile devices.

In particular, let $I$ be the input posed face image and $\Bar{I}$ be the mirror image of $I$; $f_I$ and $f_{\Bar{I}}$ be their deep features extracted by MobiFace structure, respectively. As illustrated in Fig. \ref{fig:FlippedMobiFace}, 
the Flipped-MobiFace is an extended structure of MobiFace with two additional layers that learn to predict $f_{\Bar{I}}$ from $f_I$ directly withour reruning $\Bar{I}$ through the MobiFace structure again. Then the new feature $f=\frac{1}{2} (f_I + f_{\Bar{I}})$ can be used as the final ``frontal'' features of $I$.
We employ fully-connected layers with ReLU for these extended layers.
In order to obtain the weights for these layers, for each training image $I$, we compute its flipped image $\Bar{I}$ by mirroring the face along $x$-axis. Then MobiFace is adopted to extract $f_I$ and $f_{\Bar{I}}$. Finally, the new layers are trained to obtain $f_{\Bar{I}}$ from $f_I$. The loss function consists of two main terms: (1) $\ell_2$-norm between the ground-truth $f_{\Bar{I}}$ and the predicted features; and (2) the cross-entropy loss to ensure the predicted features maintain the correct ID of the subject. 

\section{Experimental results}
We first train the network using the cleaned training set of MS-Celeb-1M \cite{guo2016ms} including 3.8 million photos from 85K subjects. Then the trained network is evaluated on two common large-scale face verification benchmarks in unconstrained environments such as Labeled Faces in the Wild (LFW) \cite{huang2008labeled}, and Megaface \cite{kemelmacher2016megaface} datasets.
This training data has no overlapping with the testing data.

The databases that are used for training and testing are first described in next subsections. Then the comparisons between different models in terms of both accuracy and model sizes are represented. MobiFace can  achieve  very  high performance, even competitive against other large-scale deep networks for face recognition.

\subsection{Databases}

\textbf{MS-Celeb-1M} \cite{guo2016ms} is introduced as a large-scale face dataset with 10 million photos of 100K celebrities. However, it also contains a large number of noisy image or wrong ID labels. To obtain a high-quality training data, the MS-Celeb-1M cleaned up the MS-Celeb-1M by computing the center feature of each subject and ranking their face images using the distance to identity center. The ones far from the center are automatically removed. Some manual checks are also employed. The refined MS-Celeb-1M consists of 3.8M photos from 85K identities.

\textbf{Labeled Faces in the Wild (LFW)} \cite{huang2008labeled} is one of the common testing dataset for face verification. LFW consists 13,233 in-the-wild facial images of 5749 subjects collected from the web. The face variations include pose, expression and illuminations. According to the testing protocol of LFW, there are 6000 face pairs where half of them are positive pairs.

\textbf{MegaFace} \cite{kemelmacher2016megaface} is one of largest publicly available testing dataset for face verification. This testing protocol is very challenging with million scale of distractors, i.e. subjects are not in the testing set. There are two main sets in Megaface, i.e. gallery and probe set. The gallery set is collected from Flickr photos and consists of more than 1 millions images from 690K identities. The probe set in Megaface are collected from two existing databases: Facescrub and FG-NET. 
As the Facescrub probe set aims at the robustness of face recognition systems on large number of identity, this set includes 100K photos of 530 subjects. Meanwhile, the FG-NET probe set focuses on the robustness of the system against age changing, with 1002 images of 82 identities from 0 to 69 years old. In this paper, we evaluate the performance of our light-weight network on the Facescrub probe set.   

\subsection{Implementation details}
In the preprocessing step, MTCNN method \cite{zhang2016joint} is applied to detect all faces and their five landmark points, i.e. two eye centers, nose and two mouth corner, in both training and testing photo.  
Then, using the information from five landmark points, each face is aligned and cropped  into a template with the size of $112 \times 112 \times 3$. This template is then normalized into $[-1,1]$ by subtracting the mean pixel value, i.e. 127.5, and divided by 128. For Flipped-MobiFace, we use two 512-dim fully-connected layers with ReLU. 

During training stage, we adopt Stochastic Gradient Descent (SGD) optimizer with the batch size of 1024. The momentum parameter is set to $0.9$. The learning rate is initialized to $0.1$ and decreases by a factor of 10 periodically at 40K, 60K, and 80K iterations. The training stage is stopped at 100K iteration.

\begin{table}[!t]
	\footnotesize
	\centering
	\caption{Performance of Different face matching methods on LFW benchmark. $^*$ stands for our re-implementation. The inference time is measured on an Intel Core i7-6850K CPU @3.6GHz.}
	\label{tb:LFW_test} 
	\begin{tabular}{|l|c|c|c|c|}
		\hline
		\textbf{Methods}  & \begin{tabular}{@{}c@{}} \# \\ \textbf{Training} \\ \textbf{images}\end{tabular} & \begin{tabular}{@{}c@{}}\textbf{Model}\\\textbf{Size}\end{tabular}&
		\begin{tabular}{@{}c@{}}\textbf{Speed}\\\textbf{(ms)} \end{tabular}& \begin{tabular}{@{}c@{}}\textbf{Accuracy} \\(\%)\end{tabular} \\  
		\hline \hline
		Google-FaceNet \cite{schroff2015facenet} & 200M & 30MB& $-$& 99.63\%\\
		CosFace \cite{Wang_2018_CVPR}& 5M&$-$ &$-$& 99.73\%\\
		LightCNN \cite{wu2018light}& 4M & 50MB &$-$& 99.33\%\\
		\hline \hline
		MobilenetV1 \cite{Howard-MobileNet} $^*$ & 3.8M&112MB & 56ms & 99.50\%\\
		MobileFaceNet \cite{chen2018MobileFaceNets} $^*$ & 3.8M & 4MB& 30ms& 99.48\%\\
		\textbf{MobiFace} & 3.8M & 9.3MB & 26ms& 99.72\% \\
		\textbf{Flipped-MobiFace} & 3.8M & 11.3MB &28ms& \textbf{99.73\%} \\
		\hline
	\end{tabular}
\end{table}

\subsection{Face Verification accuracy}
\textbf{LFW benchmark.} We first compare our MobiFace against many existing face recognition approaches including both large-scale deep models and small-scale one. Table \ref{tb:LFW_test} represented the performance of different matching methods on LFW benchmark. From these results, one can easily see that our MobiFace achieves $99.72\%$ with the model size of only 9.3MB. With this performance, our MobiFace outperforms other small-size models and achieves competitive results to other large-scale deep models. MobiFace also outperforms most of other approaches in Table \ref{tb:LFW_test}.

\textbf{Megaface benchmark.} We further validate the performance of our light-weight MobiFace on the challenging Megaface benchmark against millions of distractors. 
Table \ref{tb:Megaface_test} illustrates the verification results of different methods on Megaface. The accuracy is reported on the True Accepted Rate (TAR) at the False Accepted Rate (FAR) of $10^{-6}$. These results again emphasize the performance of our MobiFace when it outperforms the other light-weight MobileFaceNet model. Compared to other large-scale deep networks, our MobiFace has the advantages of both comparable performance to these models while maintaining low computational cost. Therefore, our MobiFace is easy to be deployed on mobile devices.

\begin{table}[!t]
	\small
	\centering
	\caption{Performance of Different face matching methods on the refined version of Megaface benchmark with one million distractors. All the models are train witn Large training dataset, i.e. $>$ 0.5M. Model size is large when its size is greater than 20MB. $^*$ stands for our re-implementation.}
	\label{tb:Megaface_test} 
	\begin{tabular}{|l|c|c|}
		\hline
		\textbf{Methods}   & \begin{tabular}{@{}c@{}}\textbf{Model }\\\textbf{Size}\end{tabular}& \begin{tabular}{@{}c@{}}\textbf{Accuracy} \\(\%)\end{tabular} \\  
		\hline \hline
		MobilenetV1 \cite{Howard-MobileNet} $^*$  & Large & 92.65\%\\
		Google-FaceNet \cite{schroff2015facenet} & Large & 86.47\%\\
		\hline \hline
		MobileFaceNet \cite{chen2018MobileFaceNets}$^*$  & Small& 90.71\%\\
		\textbf{MobiFace (Ours)}  & Small & \textbf{90.9\%} \\
		\textbf{Flipped-MobiFace (Ours)} & Small& \textbf{91.3\%} \\
		\hline
	\end{tabular}
\end{table}

\section{Conclusion}
This paper has reviewed different lightweight deep network structures and approaches for mobile devices where the computational resource is very limited. Inspired by different network design strategies, this paper has further presented a novel simple but high-performance deep network for face recognition, named MobiFace. Experiments on two common large-scale face verification benchmarks with photo in unconstrained environment have shown the efficiency of our MobiFace in terms of both accuracy and small model size. Although the model is very small, its performance on both testing benchmarks is competitive against other large-scale deep face recognition network.

{\small
\bibliographystyle{ieee}
\bibliography{submission_example}
}

\end{document}